\documentclass[journal]{IEEEtran}

\usepackage[linesnumbered,ruled]{algorithm2e}
\usepackage{amssymb}
\usepackage{amsmath}
\usepackage{caption}
\usepackage{xcolor}
\usepackage[pdftex]{graphicx}
\usepackage{graphics}
\usepackage[caption=false,font=footnotesize]{subfig}
\usepackage{cite}
\ifCLASSINFOpdf
   \usepackage[pdftex]{graphicx}
\else  
   \usepackage[dvips]{graphicx}  
\fi

\usepackage[linesnumbered,ruled]{algorithm2e}
\usepackage{amssymb}
\usepackage{xcolor}

\begin{document}
%
\title{Lateralized Learning for Multi-Class Visual Classification Tasks}

\author{Abubakar~Siddique,~\IEEEmembership{Member,~IEEE,}
        Will~N.~Browne,~\IEEEmembership{Member,~IEEE}
        and~Gina~M.~Grimshaw
\thanks{Dr. Siddique is with the School of Engineering and Computer Science, Victoria University of Wellington, PO Box 600, Wellington 6140, New Zealand.}
\thanks{Prof. Browne is with the Faculty of Engineering, School of Electrical Engineering \& Robotics, Queensland University of Technology, Brisbane 4000, Australia.}
\thanks{A/Prof. Grimshaw is with the Cognitive and Affective Neuroscience lab at the School of Psychology, Victoria University of Wellington, PO Box 600, Wellington 6140, New Zealand.}
\thanks{Manuscript received -- --, ----; revised -- --, ----.}}

\markboth{}%
{Siddique \MakeLowercase{\textit{et al.}}: Lateralized Learning: Effectively Solve Multi-Class Visual Classification Tasks and Exhibit Robustness Against Adversarial Attacks}

\maketitle


\begin{abstract}
The majority of computer vision algorithms fail to find higher-order (abstract) patterns in an image so are not robust against adversarial attacks, unlike human lateralized vision. Deep learning considers each input pixel in a homogeneous manner such that different parts of a ``locality-sensitive hashing table'' are often not connected, meaning higher-order patterns are not discovered. Hence these systems are not robust against noisy, irrelevant, and redundant data, resulting in the wrong prediction being made with high confidence. Conversely, vertebrate brains afford heterogeneous knowledge representation through lateralization, enabling modular learning at different levels of abstraction. This work aims to verify the effectiveness, scalability, and robustness of a lateralized approach to real-world problems that contain noisy, irrelevant, and redundant data. The experimental results of multi-class ($200$ classes) image classification show that the novel system effectively learns knowledge representation at multiple levels of abstraction making it more robust than other state-of-the-art techniques. Crucially, the novel lateralized system outperformed all the state-of-the-art deep learning-based systems for the classification of normal and adversarial images by $19.05\% - 41.02\%$ and $1.36\% - 49.22\%$, respectively. Findings demonstrate the value of heterogeneous and lateralized learning for computer vision applications.
\end{abstract}


\IEEEpeerreviewmaketitle
\section{Introduction}
\IEEEPARstart{D}{eep} learning (DL) for computer vision (CV) applications are commonly homogeneous systems that lack robustness because they can not capture abstract patterns in data \cite{shanahan2020explicitly,siddique2021lateralizedthesis}. One reason for the poor robustness is the reliance on homogeneous knowledge representation that produces a locality-sensitive hashing table. A homogeneous system is one where each input is treated in the same manner, i.e., it contains a representation that produces the equivalent of a locality-sensitive hashing table \cite{shanahan2020explicitly,frazier2019improving}. An artificial neural network (ANN) is a prime example of a homogeneous system because the knowledge stored in a layer in ANN treats each pixel in the same way even if the connecting weights are different.

Homogeneous systems (see Section \ref{Bckgrnd}) work well when there are only simple features in the domain or when the relationship between features and target (e.g. class or action) is linearly separable, i.e. a linear combination of features can be used to separate out specific targets \cite{butz2002anticipatory,iqbal2013reusing}. However, these systems struggle when there are complex patterns of features in the domain \cite{suzuki2019hierarchical,konidaris2019necessity}. Moreover, the majority of the homogeneous systems are not robust against noisy, irrelevant, and redundant data. A single, targeted pattern can disrupt performance accuracy. For example, DL-based systems are highly vulnerable to adversarial attacks\footnote{An adversarial attack is a technique that attempts to fool AI systems by generating deceptive images \cite{moosavi2016deepfool,su2019one}.} \cite{heaven2019deep,akhtar2018threat,madry2017towards}. Even a very small (one pixel) perturbation to an image can fool many DL-based systems resulting in the wrong prediction made with high confidence \cite{moosavi2016deepfool,su2019one}.

Existing approaches to improve robustness against adversarial attacks provide only a partial solution to these problems. The adversarial training approach is an attempt to train a DL-based system with noisy, irrelevant, redundant/adversarial data. It reduces over-fitting by regularizing the deep networks. However, this approach is expensive because it requires additional training data. Moreover, this approach is considered non-adaptive due to reliance on already existing redundant/adversarial data. Furthermore, the resultant adversarial-trained systems may still not be robust against novel adversarial patterns \cite{goodfellow2014explaining,sankaranarayanan2018regularizing,moosavi2017universal}.

A hypothesized solution is to develop heterogeneous feature-based systems (see Section \ref{Bckgrnd}). The heterogeneous features are parsimonious $-$ they only encode necessary information. These features can represent knowledge at a constituent or a holistic level, e.g., a holistic pattern of a swan, as well as its constituent parts of the beak, neck, wing, and so forth. Each knowledge component has sufficient detail to solve a part of a problem or the whole problem. Moreover, heterogeneous features can form hierarchies of knowledge such that new knowledge can be constructed from existing knowledge in a recursive manner. It is hypothesized that learning of patterns within the constituent parts in \textit{relation} to the holistic classes aids classification, e.g., a swan and emu both have long necks, but they have relative differences in beak length. 

Biological intelligence supports the use of heterogeneous knowledge representation at different levels of abstraction. Two organizing principles of vertebrate brains enable the reuse of learned knowledge at multiple levels of abstraction, i.e., lateralization and modularity of function \cite{corballis2017evolution,nitz2012spaces,krichmar2008neuromodulatory}. It is hypothesized that a lateralized AI system can robustly handle uncertainty, noise, irrelevant, and redundant/adversarial data by splitting knowledge representation into constituents and holistic components.  

\begin{figure}
	\begin{center}
		\includegraphics [scale=0.27]{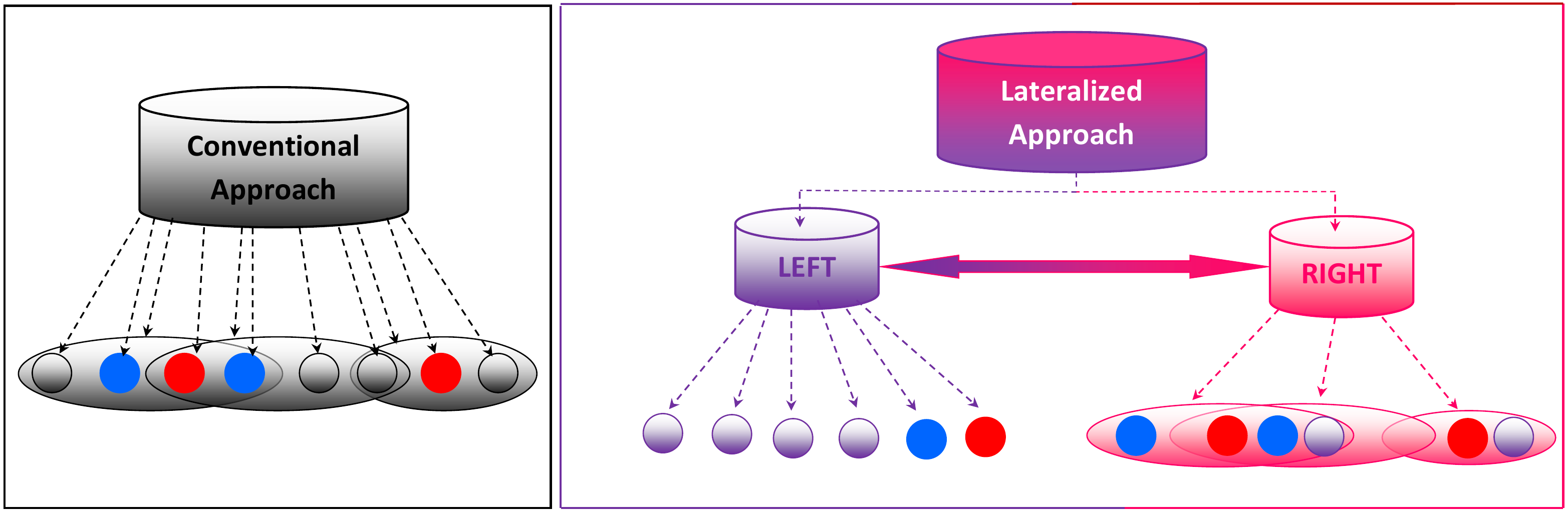}
		\vspace{-0.70em}
		\caption{A schematic illustration of a conventional homogeneous approach (A) and a lateralized approach (B). A conventional homogeneous approach considers individual features and niches in a homogeneous manner. Whereas, a lateralized approach splits a complex problem into constituent (left half) and abstract (right half) knowledge.}
		\vspace{-0.5em}
		\label{ConvSysLatSys}
	\end{center}
\end{figure}

Many real-life problems are constructed from sub-problems. For example, the classification of birds has a hierarchical taxonomy from order (e.g., songbirds), through family, the genus to species. Thus a learning system must be able to represent knowledge at different levels of abstraction. This ability enables the learning system to address constituents of the problem and higher level (big picture) simultaneously. Consequently, the robustness of the learning system against adversarial attacks is enhanced. Both the detailed sub-components and the overall pattern must be successfully challenged to fool the learning system. A schematic illustration of the contrast between a conventional homogeneous approach and a novel lateralized approach is shown in Fig. \ref{ConvSysLatSys}.

To evaluate the potential benefits of a heterogeneous approach, we previously developed a basic symbolic lateralized AI system to address binary-class image classification tasks \cite{siddique2020lateralized}. By representing an input image at different levels of abstraction, it has shown promising performance while allowing the contributions of lateralization to be interrogated. Despite the promising results, the developed system could not handle a large number of features/classes search spaces. The main reasons for the poor performance of the developed system are its symbolic nature and methods that lacked scalability. The majority of real-world problems have many classes and a large number of features. Thus a novel lateralized system for real-world tasks is needed to overcome the limitations of the basic symbolic lateralized AI system.

\subsection{Contributions}
This study creates a novel lateralized system, inspired by the principles of cognitive neuroscience, that is scalable and can handle a large number of features, classes, and search spaces. The novel system applies lateralization and modular learning at different levels of abstraction to exhibit robustness in a real-world domain that includes uncertainty, noise, and irrelevant and redundant data. The main contributions of this study are as follows:

\begin{itemize}
    \item[(i)] We devise a lateralized system that is, for the first time, scalable and can handle a large number of features, classes, and search spaces. The novel system can exhibit robustness against adversarial attacks in visual tasks.
    \item[(ii)] A novel constituents class matrix is created that can handle a large number of classes in real-world problems. The methodologies of constituent-level perception and holistic-level perception are enhanced to support multiple classes.
    \item[(iii)] We create a novel strategy to learn patterns at constituent levels (components) that aid classification at holistic levels. The proposed strategy utilizes heterogeneous features to form hierarchies of knowledge such that new knowledge can be constructed from existing knowledge in a recursive manner.
    \item[(iv)] The decision-making process of the novel lateralized system is interpretable, which is a step toward eXplainable AI.
\end{itemize}

The remainder of this paper is organized as follows: Section \ref{Bckgrnd} presents the relevant findings from cognitive neuroscience that inspire this work. It also includes the required background knowledge from machine learning and computer vision. Section \ref{LatSys} describes how a lateralized system can be created. It explains the critical components and learning mechanisms of the novel system. An example task of classifying birds ($200$ classes), in photographic images, is used to show the effectiveness and scalability of the lateralized approach. The robustness of the developed system against noisy, irrelevant, and redundant data is presented in Section \ref{ExpWork}. Section \ref{disc} highlights the strengths, limitations, and drawbacks of the lateralized approach. Finally, the Conclusion and Future work are provided in Section \ref{Con}.

\section{Background}
\label{Bckgrnd}
The goal of this section is two-fold: first, to introduce the relevant principles of cognitive neuroscience that will inform this work; and second, to provide the relevant background knowledge of computer vision and machine learning that provide a foundation upon which the novel lateralized system will be developed. 

\subsection{Cognitive Architecture in Vertebrate Brains}
\label{CogArch}
The cognitive architecture in vertebrate brains allows them to learn knowledge from simple and small-scale problems and then reuse it at different levels of abstraction to resolve complex and large-scale problems \cite{ralph2017neural,binder2016toward,ralph2014neurocognitive}. It is not the aim of this work to model the cognitive architecture of a specific species; rather takes inspiration from basic principles of vertebrate intelligence. Hemispheric lateralization is an important feature of vertebrate intelligence that is relevant to this work.

The propensity for one hemisphere to perform a specific cognitive process more precisely and efficiently than the other is called hemispheric lateralization \cite{banich2010cognitive}. Both the left and right hemispheres look-alike in the macro-structural view. However, they have distinct neuroanatomy, neurochemistry, and functional architecture \cite{banich2010cognitive}. Three key aspects of lateralization are presented below.

\subsubsection{Representation and Processing}
In vertebrate brains, some functions are entirely lateralized to one hemisphere or the other. For instance, each hemisphere receives sensory input from the other side of the body and controls the contralateral musculature. But outright lateralization is a special case, not the standard. Both the left and right hemispheres contribute to the majority of higher-order perceptual and cognitive tasks. The differences between the hemispheres are more relative than absolute. Often these hemispheric differences are related to the scale of processing the same input sensory information. For instance, in visual perception, the left hemisphere processes the input signal at a constituent (local) level while the right hemisphere processes the same input signal at a holistic (configural or global) level \cite{robertson1991neuropsychological,robertson1986part,martin1979hemispheric}.  Similarly, in speech perception, the left hemisphere processes segmental information (individual phonemes that make up words) while the right hemisphere processes super-segmental information (global intonational patterns that reflect emotion or intention of the speaker) \cite{hickok2016neural,grimshaw2009once,wildgruber2006cerebral}. 

The Double Filtering by Frequency model (DFFM) clarifies these crucial differences in the hemispheric representational scale. The DFFM proposes that the left and right hemispheres act as high-pass and low-pass filters, respectively. Consequently, the left hemisphere represents the detailed information that is available only in high temporal or spatial frequencies of the input signal; the right hemisphere represents global patterns that are available only in low temporal or spatial frequencies \cite{flevaris2016spatial,robertson2000hemispheric}. Such complementary representations are not limited to sensory processing only. For instance, in language processing, the literal meanings of words and sentences are activated by the left hemisphere; while the alternative, figurative, metaphorical meanings are activated by the right hemisphere \cite{faust2014rigidity}. This ability to represent the same input signal (problem instance) at a constituent (local) level and a holistic (configural or global) level will need to be incorporated in the work presented here (Obj. i).

\subsubsection{Coordination}
The computations carried out in opposite hemispheres need to be coordinated for effective cognition. For instance, recognizing a face requires us to integrate elementary features (left) with their configural arrangements (right) \cite{frassle2016mechanisms}; understanding a joke requires us to integrate the literal meaning of words and sentences (left) with their alternative subtext (right) \cite{chan2013towards}; understanding a song requires us to integrate the lyrics (left) with the melody (right) \cite{yelle2009hemispheric}. It is the coordination between the left and right hemispheres that allows the transfer of critical information at different levels of abstraction. This coordination needs to be incorporated into the system modules created here (Obj. ii).

\subsubsection{Goal-driven Processing}
The vertebrate brain has the ability to select the most appropriate and relevant hemisphere to perform the computations required for a specific task. This selection is managed by the goal-driven processes that analyze the problem instance and shift control to the appropriate and/or superior hemisphere \cite{dharmaretnam2005hemispheric,rogers2004advantages}. For example, in language processing, the right hemispheres' speech processing is prioritized if the emotional state of a conversational partner is more relevant; however, the left hemispheres' elemental processing is dominated if the linguistic elements are of concern \cite{dharmaretnam2005hemispheric,rogers2004advantages}. Depending on the nature of the task, the connections between the left and right hemispheres can be inhibitory or excitatory to enable inhibition or integration, respectively \cite{stark2008regional}. The ability to identify which hemisphere is best for the processing of a specific task is important in real-world problems. A strategy needs to be developed for the identification/activation of the most suitable system module concerning the problem at hand (Obj. iii).

\subsection{Computer Vision}
This section presents a brief introduction to the associated computer vision techniques (i.e. image features and adversarial attack) that are used in this work. Often, image features have been used as input instances for the classifiers to label the image. A feature can be considered as a piece of information about the visual contents of an image. The number of features is few as compared with a large amount of image data (raw pixels of an image) \cite{siddique2016comprehensive,al2017genetic}. Adversarial attacks have been used to generate noisy, redundant, and irrelevant data to corrupt images \cite{goodfellow2014explaining,madry2017towards,kurakin2016adversariale}. The resultant images can be used to evaluate the robustness of AI systems.

\subsubsection{Features}
\label{Feat}
Scale-invariant feature transformation (SIFT) is a state-of-the-art technique that has been commonly used for object detection and image classification \cite{lowe2004distinctive}. The SIFT descriptor computes receptive fields based on image measurements. These receptive fields are used to create local scale-invariant reference frames. The resultant SIFT features are invariant to illumination changes, uniform scaling, and orientation. Thus the SIFT features will assist the novel system to exhibit robustness against occlusion, noise, and/or disruption by clutter \cite{lowe1999object,ke2004efficient}.

Histogram of oriented gradients (HOG) is another well-recognized and commonly used technique for image classification \cite{dalal2005histograms}. The HOG descriptor computes a pixel-wise histogram of gradient directions. The resultant HOG features are invariant to light conditions, geometric transformation, and/or color variation. Moreover, a HOG descriptor can accurately detect object deformation and complex shapes based on the distribution of local gradient, e.g. facial expression and muscle movements \cite{chen2014facial}. Thus the HOG features will assist the novel lateralized system to exhibit robustness against noisy, irrelevant, and redundant data.

\subsubsection{Adversarial Attacks}
An adversarial attack is a technique that attempts to fool AI systems by generating deceptive input \cite{moosavi2016deepfool,su2019one}. The fast gradient sign method (FGSM) is one of the widely used and well-recognized techniques to generate adversarial images. Homogeneous systems, especially deep networks, encourage linear behavior for efficient learning. However, it has been observed that the adversarial images generated by the FGSM technique exploit this linearity of AI systems in higher-dimensional space. Moreover, the FGSM technique creates perturbations, by performing a one-step gradient update at each pixel, with the aim to increase the loss of a classifier \cite{goodfellow2014explaining}. 

In many real-world applications, AI systems receive input data through different devices such as cameras and sensors. The iterative adversarial technique has been proposed to generate adversarial images for real-world applications \cite{kurakin2016adversariale}. Instead of applying a one-step gradient update only, the iterative technique enhances the FGSM approach and makes finer optimizations (small changes) in multiple iterations. The adversarial images generated by these techniques will be used to evaluate the robustness of the novel lateralized system. 

\subsection{Machine Learning}
This section presents an overview of two important techniques for image classification, i.e. deep learning and random forest. Moreover, it explains the important differences between homogeneous and heterogeneous systems w.r.t. this work. Furthermore, it describes our previous work to create lateralized systems.

\subsubsection{Deep Learning}
Deep learning (DL) is a well-known methodology that has been used to extract useful patterns and higher-level features from raw data \cite{dalal2005histograms,ke2004efficient}. DL-based systems represent knowledge in the form of artificial neural networks. The connections between the nodes of a network have weights. These weights are updated, by utilizing methods such as backpropagation, to support the desired learning of input to output relationships. DL-based systems use multiple layers of neural networks to extract useful information. Each previous layer's signal is transformed into a subsequent filtered representation. Consequently, the DL-based system progressively models the patterns in the given problem. This chain of information transformation from the input layer to the output layer is controlled through the credit assignment path (CAP). In feedforward neural networks, the CAP consists of the number of hidden layers plus the final output layer, whereas, in recurrent neural networks it is unlimited because a signal may propagate through a layer multiple times \cite{schmidhuber2015deep}. Moreover, DL-based systems generate a homogeneous knowledge representation, such that all features are treated equally in each layer, to learn a linearly separable relationship between features and target \cite{madry2017towards,goodfellow2014explaining,kurakin2016adversariale}. In contrast, the novel lateralized system will represent the same input signal at a constituent-level and holistic-level simultaneously. This is anticipated to enable robustness against noisy, irrelevant, and redundant data.

\subsubsection{Random Forest}
Random Forest (RF) is a state-of-the-art technique that can directly handle high-dimensional data sets. RFs have been commonly used for regression and classification (binary/multi-class) problems \cite{ho1998random,breiman2001random,zhang2012ensemble}. RF is a decision tree-based ensemble learning method that works by constructing multiple decision tree (DT) classifiers. It makes the final prediction based on the votes of the decision trees, i.e. the class that is the mode of the outputs of these DTs \cite{ho1998random}. The RF algorithm starts by selecting many bootstrap samples from the data. The instances that are not selected in bootstrap are called out‐of‐bag instances. A DT classifier is constructed for each bootstrap such that at each node in the DT, only a subspace of randomly selected features is considered for the split process. These DT classifiers are used to predict out-of-bag instances. The final prediction class for an instance is obtained by the majority vote of the out‐of‐bag predictions for that instance \cite{breiman2001random}. 

\subsubsection{Homogeneous System}
The prefix ``homo'' is derived from Greek, meaning ``same''. The knowledge that is represented (stored) in the same manner can be referred to as homogeneous knowledge. For example, the knowledge stored in a layer in artificial neural networks treats each element in the same way even if the connecting weights are different. In this work, an AI system is called a homogeneous system if it equally considers all the features of an input environmental signal in the same manner. A deep network with a convolution filter and a single pathway stride is an obvious example of a homogeneous system. Similarly, a deep network-based system with different convolution filters (e.g. $3\times3, 5\times5$) and different pathway strides, which are combined in the end in an ensemble way, is still a homogeneous system. 

\subsubsection{Heterogeneous Systems}:
The word ``hetero'' is also a Greek prefix, which means ``different''. The knowledge that is represented (stored) at different levels of abstraction can be referred to as heterogeneous knowledge. In this work, an AI system is called a heterogeneous system if it has the ability to consider an input environmental signal at different levels of abstraction\footnote{A level of abstraction can be considered as a perspective to address a problem. For example, in a visual task, a local viewpoint (eyes, nose, and mouth of a cat) and big-picture (whole cat) are two different levels of abstraction.} simultaneously to generate a heterogeneous knowledge representation. A deep network with different convolution filters (e.g. $3\times3, 5\times5$) and different pathway strides, which are applied at different parts of the same image such that they can share their findings within the hidden layers can be considered as a heterogeneous system. Capsule networks \cite{sabour2017dynamic} is a step toward creating heterogeneous neural networks as they extract different spatial relationships of features within a single layer. Holistic level features are generally heterogeneous as they may entail hierarchical patterns within patterns. Heterogeneous features can be parsimonious as they only encode necessary information. 

\subsubsection{Previous Work on Lateralized Systems}
\label{PrevWorkLatSys}
We developed different lateralized systems to solve complex problems in different domains such as Boolean, computer vision, and navigation \cite{siddique2021lateralizedthesis}. Initially, we developed a lateralized AI system for complex Boolean problems to obtain a proof-of-concept of the novel lateralized approach \cite{siddique2022lateralized}. The novel system applies reinforcement learning techniques to solve interogatable, single-step, scalable, and complex Boolean problems (all features and classes take value $\{0,1\}$). Boolean problems are complex engineering problems but these problems are neither real-world nor have uncertainty.

The next stage was to consider lateralization for temporal problems (prior to the work here on multi-class, real-world problems). The majority of homogeneous systems struggle to capture complex structures in an environment spatially (in single-step problem domains) but also temporally (in multi-step domains). For example, perceptual aliasing is a long-standing problem for artificial agents in applying reinforcement learning (RL) to many multi-step tasks \cite{frazier2019improving,suzuki2019hierarchical,whitehead1991learning,chrisman1992reinforcement,lanzi2000adaptive,butz2002anticipatory,zatuchna2005agentp}. A novel frame-of-reference (FoR) based lateralized AI system was created for navigation problems to evaluate the effectiveness of the lateralized approach \cite{siddique2021frames}. The novel system applies a reinforcement learning technique to address perceptual aliasing in multi-step decision making tasks.

The next step to evaluate the effectiveness of the lateralized approach could be evaluated is in a real-world domain with uncertainty, noise, irrelevant, and redundant data. The majority of AI systems do not exhibit robustness against noisy and irrelevant data. For example, deep artificial neural networks are highly vulnerable to adversarial attacks in visual classification tasks \cite{akhtar2018threat,madry2017towards}. We developed the basic symbolic lateralized AI system to address binary-class image classification tasks \cite{siddique2020lateralized}. This system utilized symbols to aid the interpretability of the discovered knowledge. However, this restricted scaling in problems with a large number of classes due to the multi-combinatorial capacity of symbols leading to intractable search space. By considering an input image at different levels of abstraction, the basic lateralized system has shown promising performance while allowing the contribution of lateralization to be interrogated. 

Despite the promising results, there were major concerns that all these previous lateralized systems have been developed by using the underlying technique of learning classifier systems (LCSs). LCSs based systems, due to their symbolic nature, would struggle in a real-world large number of features/classes search spaces. Moreover, the basic lateralized system developed for the cat vs dog dataset was limited to binary-class image classification tasks. This work aims to develop a novel lateralized system to show that the lateralized approach can be scaled, adapted to other real-world problems, and not limited to LCSs.

\section{Lateralized System}
\label{LatSys}
The novel lateralized system enhances/changes the architecture of the basic lateralized system from the following aspects: (i) Instead of binary classes, a constituents class matrix is implemented that can handle any number of classes. Subsequently, the methodologies of constituent-level perception and holistic-level perception are enhanced to handle multiple classes; (ii) The underlying technique of the learning classifier system is replaced by the random forest to address high-dimensional search spaces. (iii) The constituents level prediction components are increased from $3$ to $11$ to provide a better insight into the input image.

\begin{figure}
	\begin{center}
	    \includegraphics [scale=0.35]{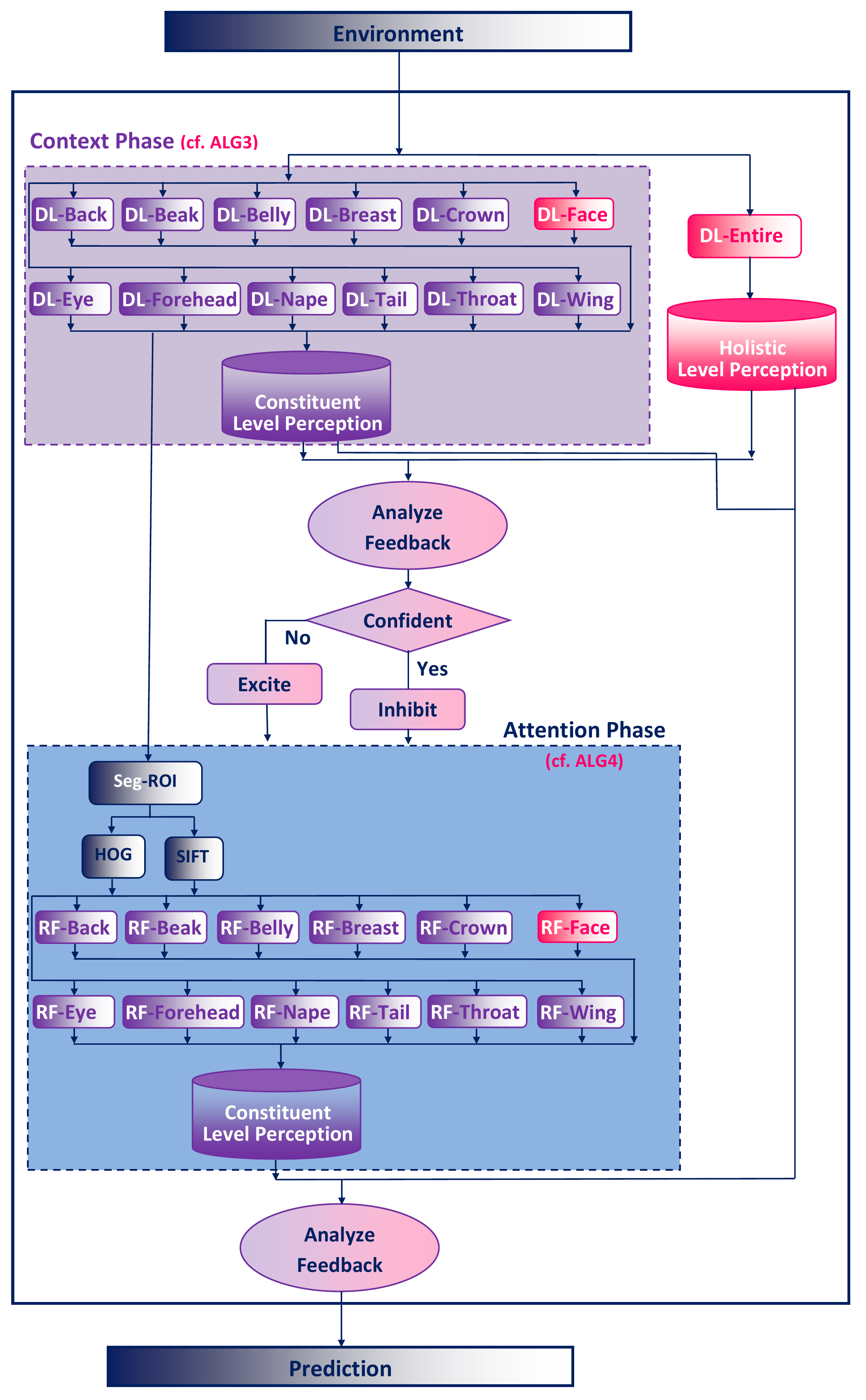}
		\caption{A schematic depiction of the strategies developed to achieve cognitive inspired functionality in the multi-class lateralized system. (DL: deep learning, RF: random forest)}
		\label{FlwChrtMulti}
	\end{center}
\end{figure}

The overall classification process of the novel system is described first, prior to explaining the details. The lateralized system can be divided into two main phases, i.e. the context phase and the attention phase. The context phase handles simple images, whereas, the attention phase addresses noisy and corrupt images. The novel lateralized system is created by adapting the lateralized approach such that both phases (context phase and attention phase) implement left-half (constituent level) and right-half (holistic level) functionality. The implementation of this functionality for both phases enables lateralization at multiple levels of abstraction. Moreover, the excite and inhibit signals are generated by the context phase which assists the novel system to avoid extraneous computations and efficiently solve the given problem. Finally, the feedback from both phases is analyzed to resolve the problem. A schematic depiction of the strategies developed for the novel system is shown in Fig. \ref{FlwChrtMulti}.

\subsubsection{Context Phase}
The context phase uses DL-based models to generate constituent level (i.e., individual parts) and holistic level (i.e., big picture) predictions. The context phase consists of total $d (d = d_{1} + d_{2})$ number of deep networks such that $d_{1}$ networks are used to generate predictions about the constituents and $d_{2}$ networks are used to generate higher level configural predictions. Here, the prediction is the probability that a part/image belongs to a candidate species (class). 

For this work, a multi-class ($200$ classes) birds classification is used as an illustrative example. The context phase consists of thirteen deep networks. Eleven of these deep networks are used to generate predictions about the constituents, i.e. back, beak, belly, breast, crown, eye, forehead, nape, tail, throat, and wing; whereas two deep networks are used to generate configural predictions about the face and overall image.

Let $\mathcal{CM}_{c}$ is a constituent class matrix that contains the prediction values, i.e., the probability of an image belonging to a class. It has $n$ entries, where $n$ is a total number of classes, such that one entry for each class. Moreover, the prediction values are initialized to $0$. Each constituent deep model generates a prediction value for the given image, as shown in Equation \ref{EqM}.

\begin{equation}
\label{EqM}
    \mathcal{M}(img) =
    \begin{cases}
      C, & \text{class} \\
      P, & \text{probability}
    \end{cases}
\end{equation}

where $\mathcal{M}$ is a model that predicts class $C$ with probability $P$ for a given image $img$. These prediction values are added in the constituent class matrix, as shown in Equation \ref{EqCM}.

\begin{equation}
\label{EqCM}
    \mathcal{CM}_{c} =\sum\limits_{i=1}^n \sum\limits_{\forall m} \mathcal{I}\times P
\end{equation}

where $n$ is the number of classes (e.g., $200$), $m$ is the number of constituent models, and $\mathcal{I}$ is given below.

\begin{equation}
\label{EqI}
    \mathcal{I} =
    \begin{cases}
      1, & \text{C = i} \\
      0, & \text{Otherwise}
    \end{cases}
\end{equation}

All the entries in the $\mathcal{CM}_{c}$ are normalized between $0$ and $100$. Finally, the probability of each class in $\mathcal{CM}_{c}$ is compared and the class with the highest probability is considered as a constituent-level perception (CLP), Equation \ref{EqCLP}. 

\begin{equation}
\label{EqCLP}
\mathcal{CLP} = \max_{i \in [1,\dots, n]}\mathcal{CM}_{c}(i) 
\end{equation}

In contrast, a deep model is used to obtain the holistic-level prediction probabilities by utilizing the whole image. The resultant highest prediction probability is considered as a holistic-level perception (HLP), Equation \ref{EqHLP}. 

\begin{equation}
\label{EqHLP}
\mathcal{HLP} = \max_{i \in [1,\dots, n]}P(i) 
\end{equation}

where $P$ is the prediction probability for class $i$, and $n$ is the number of classes (e.g., $200$). It is important to note here that if a deep network is unable to predict any part (due to noise or adversarial attack), the default prediction value of $0$ is used. The $\mathcal{CLP}$ and $\mathcal{HLP}$ information are shared with the main module of the system. 

The main module analyzes the feedback from $\mathcal{CLP}$ and $\mathcal{HLP}$. If $\mathcal{CLP}$ and $\mathcal{HLP}$ support each other, i.e. both predict the same class, the main module marks the confident flag as \textit{True}, which indicates that the system can confidently classify the given image. Subsequently, the main module generates an inhibit signal to the attention phase so that it can cease working. However, if $\mathcal{CLP}$ and $\mathcal{HLP}$ do not support each other (i.e. $\mathcal{CLP}$ and $\mathcal{HLP}$ predict different classes) or $\mathcal{CLP}$ is confused (i.e. more than one class has the maximum prediction value), the main module marks the confident flag as \textit{False}, which indicates that the system cannot confidently classify the given image. Subsequently, the main module generates an excite signal to the attention phase and waits for the reply. The pseudo-code of the strategy adopted by the context phase to generate a prediction is given in Algorithm \ref{StrAlgoCntxtMulti}.

\begin{algorithm}
\LinesNumbered
\SetKwBlock{Begin}{Begin}{}
	\SetAlgoLined
	\KwData{The data set and problem configurations}
	\KwResult{Generate a prediction, inhibit or excite signal for attention phase}
	Initialize the global variable and parameter settings\;
	Compute CCM(); \qquad\textit{\% Compute Constituent Class Matrix (CCM). }\\
	\Indp Initialize CCM(); \qquad\textit{\% Initialize all the classes with $0$ probability. }\\
	Generate Prediction From Deep Model(); \qquad\textit{\% Generate prediction probability for each part (e.g., back, beak, belly, breast, crown, eye, forehead, nape, tail, throat, wing, face) or whole image. }\\
	Update CCM for each Deep Model(); \qquad\textit{\% Add all the prediction probabilities in the CCM.}\\
	Normalize CCM(); \qquad\textit{\% Normalize all the prediction values in the CCM between $0$ and $100$.}\\
	\Indm 
	Compute CLP(); \qquad\textit{\% Compare the prediction probabilities of all the classes in the CCM. The class with highest prediction probability is considered as a constituent-level perception (CLP). Note that the prediction probability for the unrecognized parts (e.g. deep network is unable to identify due to noise or adversarial attack) is $0$ by-default.}\\
	Compute HLP(); \qquad\textit{\% The highest whole image prediction probability is considered as a holistic-level perception (HLP).}\\
	Analyse Feedback ();  \qquad\textit{\%Analyze the feedback from CLP and HLP.}\\
	\Indp \eIf{CLP and HLP Predict the Same Class}{  \textit{\%The main module is confident that the system can correctly classify the given image.}\\
		Generate Inhibit Signal(); \qquad\textit{\% Generates an inhibit signal to the attention phase so that it stops.}\\
		Generate Final Prediction()\;
	}{
	    Generate Excite Signal(); \qquad\textit{\% Generates an excite signal to the attention phase.}\\
	}
	\Indm 
	\caption{Strategy adopted by the context phase to generate a prediction (cf. Fig. \ref{FlwChrtMulti}).}
	\label{StrAlgoCntxtMulti}
\end{algorithm}

\subsubsection{Attention Phase}
The attention phase uses RF-based classifiers to generate constituent-level (i.e., individual parts) and holistic-level (i.e., big picture) predictions. It consists of total $r (r = r_{1} + r_{2})$ number of random forest classifiers. The $r_{1}$ RF classifiers are used to generate predictions for individual constituent parts of the image. It is similar to the class identification in many image data sets, e.g., ImageNet. These components require human-labeled ground truths. The remaining $r-{2}$ RF classifiers are used to generate holistic-level predictions. 

In the example case disclosed here, the attention phase consists of thirteen RF classifiers. Eleven of these RFs are used to generate predictions about the constituents, i.e., back, beak, belly, breast, crown, eye, forehead, nape, tail, throat, and wing; whereas one RF is used to generate configural predictions about the face. The attention phase starts processing in parallel with the context phase. However, it stops immediately if it receives an inhibit signal from the main module. The attention phase utilizes the deep models of the context phase to generate predictions about the bounding box (bbox) of the back, beak, belly, breast, crown, eye, forehead, nape, tail, throat, wing, and face.\footnote{Separate routines are developed to create segmented images. Further explanation of these routines is not presented as independent of the lateralized approach.} Subsequently, it segments each part according to the bbox values. 

These segmented images are used to compute different types of features that form the input instances (environment) for the RF classifiers. The feature types are selected based on the nature of the problem, the number of classes, the effectiveness of the classification, and robustness against noisy, irrelevant, and redundant data. For this work, three variants of SIFT features and three variants of HOG features are computed from the segmented images (see Section \ref{Bckgrnd}). The RF classifiers utilized these features to generate prediction probabilities for each class. Another class matrix $\mathcal{CM}_{a}$ for attention phase is computed based on these probabilities by using equations \ref{EqM} and \ref{EqCM}. All the entries in the $\mathcal{CM}_{a}$ are normalized between $0$ and $100$. Finally, the $\mathcal{CLP}$ for the attention phase is computed by using equation \ref{EqCLP}. This final RF-based $\mathcal{CLP}$ value is returned to the main module of the system. 

The main module analyzes the deep model based $\mathcal{CLP}$, RF-based $\mathcal{CLP}$, and $\mathcal{HLP}$. If two of these perceptions support the same class, the main module favors that class and makes a prediction. Otherwise, the system computes another class matrix $\mathcal{CM}_{f}$ by adding the respective prediction probabilities of $\mathcal{CM}_{c}$, $\mathcal{CM}_{a}$, and holistic-level prediction probabilities. Finally, the main module compares the prediction probabilities of all the classes in $\mathcal{CM}_{f}$ and the resultant highest prediction probability is considered as the final prediction $\mathcal{FP}$, as shown in Equation \ref{EqFP}. The pseudo-code of the strategy adopted by the attention phase to generate a prediction is given in Algorithm \ref{StrAlgoAttentionMulti}.

\begin{equation}
\label{EqFP}
\mathcal{FP} = \max_{i \in [1,\dots, n]}\mathcal{CM}_{f}(i) 
\end{equation}

\begin{algorithm}
	\SetAlgoLined
	\KwData{The data set and problem configurations}
	\KwResult{Generate final prediction}
	Initialize the global variable and parameter settings\;
	Check Inhibit Signal(); \qquad\textit{\% Receives inhibit signal form the main module and stops immediately.}\\
	Get BBox From Deep Models(); \qquad\textit{\% Get bbox prediction for each part (e.g., back, beak, belly, breast, crown, eye, forehead, nape, tail, throat, wing, face).}\\
	Crop Img();  \qquad\textit{\% Segment each part based on the bbox values.}\\
	Compute SIFT(); \qquad\textit{\% Compute three variants of SIFT features for each segmented image.}\\
	Compute HOG(); \qquad\textit{\% Compute three variants of HOG features for each segmented image.}\\
	Get RF Prediction(); \qquad\textit{\% Get RF predictions for each part}\\
	Compute RF-CCM(); \qquad\textit{\% Compute RF based Constituent Class Matrix (CCM). }\\
	Compute RF-CLP();\\
	Share RF-CLP();\\
	Analyse Feedback ();  \qquad\textit{\%Analyze the feedback from RF-CLP, Context-CLP and Context-HLP.}\\
	\Indp \eIf{Majority Favor Same Class }{
	    Generate Prediction()\;
	}{
	    Compute FCM()\; \qquad\textit{\%Compute final prediction class matrix by adding the respective entries from CCM, RF-CCM, and holistic probabilities.}\\
	    Compute FP(); \qquad\textit{\% The highest prediction probability in the FCM is considered as a final prediction.}
	}
	\caption{Strategy adopted by the attention phase to generate a prediction (cf. Fig. \ref{FlwChrtMulti}).}
	\label{StrAlgoAttentionMulti}
\end{algorithm}

\section{Experimental Work}
\label{ExpWork}
This work seeks to show the scalability and robustness of the lateralized approach. Here classification experiments on a multi-class image data set, that also contains noisy, irrelevant, and redundant data, are conducted. The data set needs to have constituent-level as well as holistic-level ground truth information as it is required for the training of the constituent-level prediction models and holistic-level prediction models of the novel system. Note that such publicly available data sets are currently rare. For this work, the experiments are conducted for bird species classification. The selected data set is publicly available and it has constituents and holistic levels labeled data. Future work will consider how constituent parts could be identified automatically.

\subsection{Data Sets}
\begin{figure}
	\includegraphics[scale=0.2]{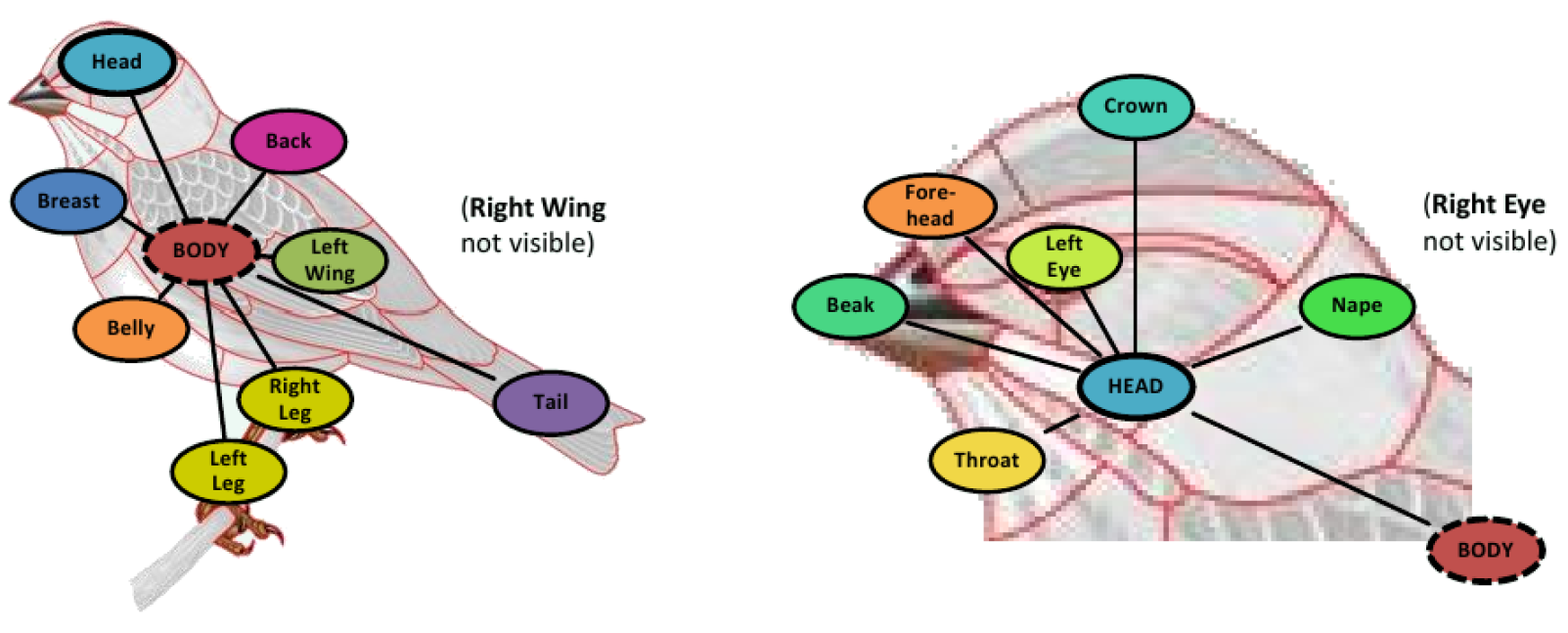}
	\caption{A sample bird image with $15$ points annotation of the parts (source \cite{wah2011caltech}).}
	\label{SampleImg-Bird}
\end{figure}

This work uses a publicly available bird data set (Caltech-UCSD Birds-200-2011) that has been used by the research community \cite{wah2011caltech}. This dataset contains $11788$ photographic images of $200$ bird species. The ground-truth information about the parts and the overall image is available. Each image contains $15$ points annotation of the bird, i.e. (1) back, (2) beak, (3) belly, (4) breast, (5) crown, (6) forehead, (7) left eye, (8) left leg, (9) left wing, (10) nape, (11) right eye, (12) right leg, (13) right wing, (14) tail, (15) throat (see Fig. \ref{SampleImg-Bird}).

\subsubsection{Data Preparation}
The lateralized system needs the information (bbox) about the back, beak, belly, breast, crown, eye, forehead, nape, tail, throat, wing, and face of a bird in an image to train deep models and RF classifiers. The ground truth files of the chosen data sets have annotations related to these parts but not the bboxs. The novel system utilizes these annotations to generate the required information. For this purpose, separate routines are developed that take the annotation file as input and generate corresponding bboxs for each part and face. Moreover, these routines have the ability to handle rotated images and generate bboxs accordingly. Further explanation (logical description and algorithm) of these routines is not presented as independent of the lateralization approach.

\subsection{Experimental Setup}
The learning strategy for the context phase is developed by utilizing multiple deep models depending on the nature and complexity of the problem, e.g. twelve deep models are used here. These models can be based on any state-of-the-art pre-trained deep network as a basis for transfer learning. The models used here are based on pre-trained $50$ layers residual networks (ResNet50 model), which are well-recognized and widely used deep networks \cite{he2016deep}. The loss function mean absolute error (MAE), with the Adam optimization algorithm (stochastic gradient descent optimizer), is used to update the network weights \cite{kingma2014adam}. The deep models are trained for $300$ epochs and the length of each epoch is $1000$ training instances. 

It is important to note that ResNet50 model is also used to generate adversarial images to ensure that the adversarial attacks have the same underlying model. The learning strategy for the attention phase is developed by utilizing multiple (e.g. twelve here) RF classifiers. The RF classifiers are used with their default settings, i.e, no parameter tuning is performed. Moreover, the system is implemented using Scikit-learn libraries version 0.21.3 for Python version 3.7 \cite{pedregosa2011scikit}.

\begin{figure*}
    \vspace{-1em}
	\subfloat[\vspace{-1em}FGSM-M based adverasarial images.\label{ExpAdvImgs-FGSM-M}]{\includegraphics[scale=0.5]{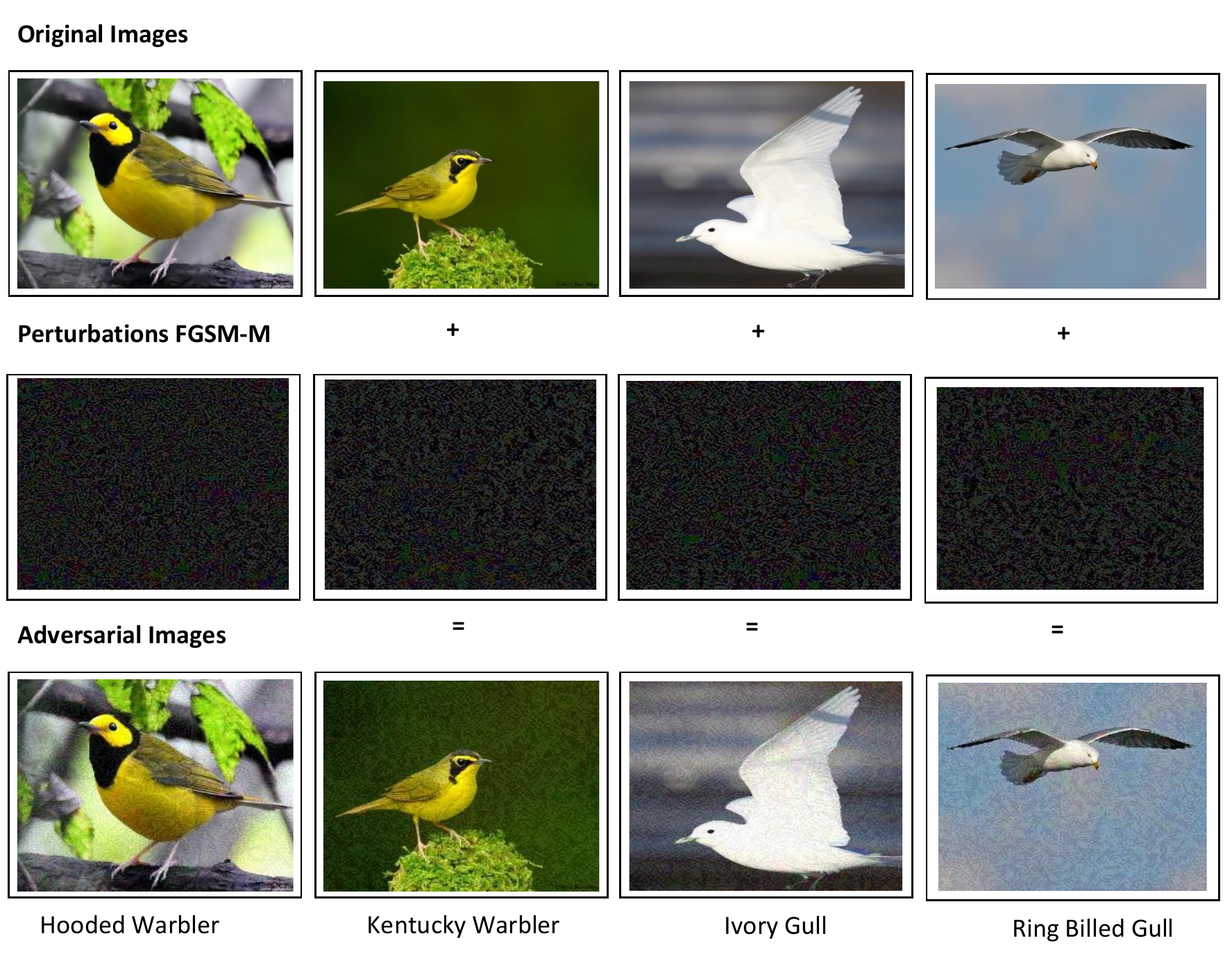}}
        \subfloat[\vspace{-1em}Itr-M based adverasarial images.\label{ExpAdvImgs-Itr-M}]{\includegraphics[scale=0.5]{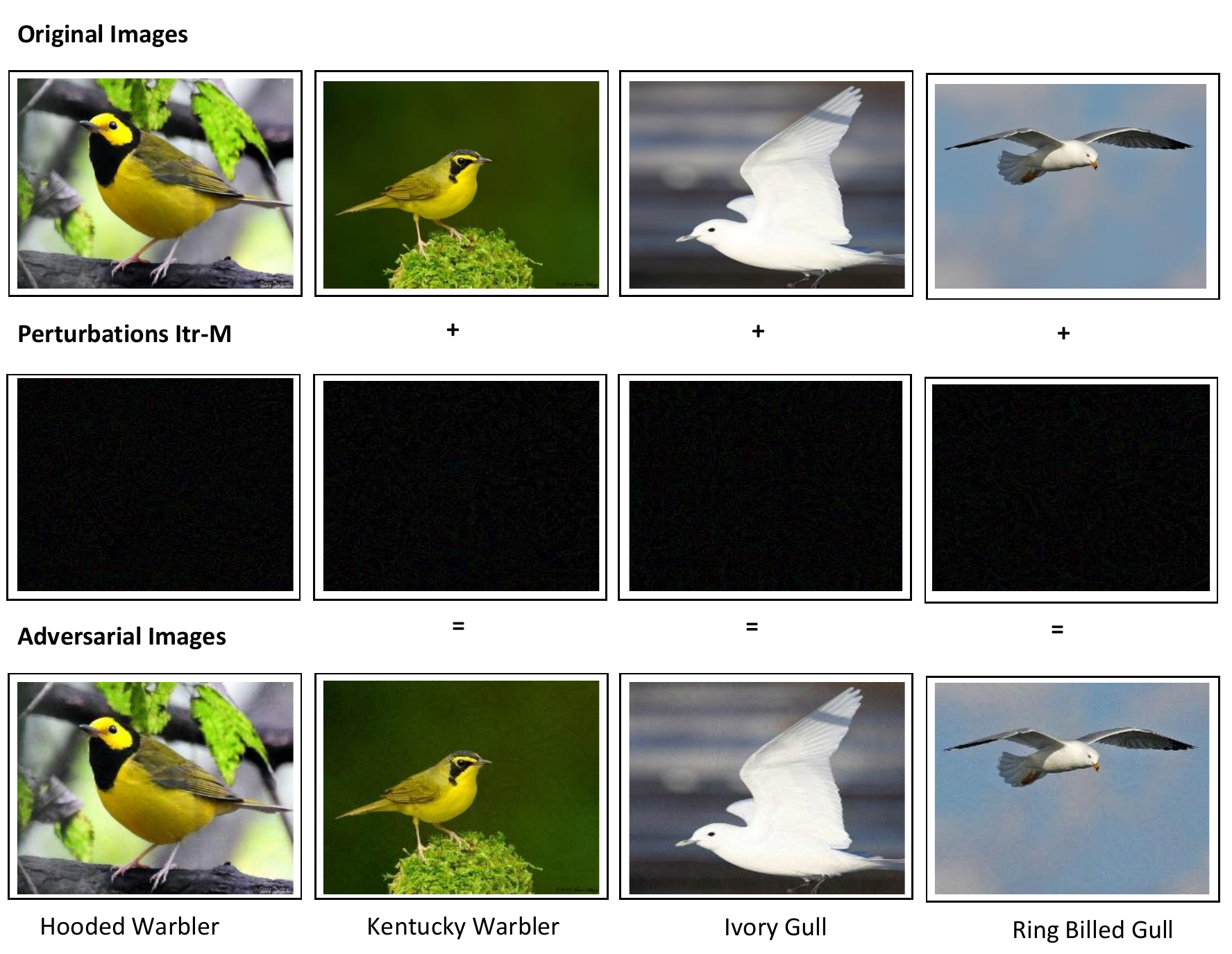}}\hfill
	\vspace{-1em}
	\subfloat[\vspace{-1em}FGSM-S based adversarial images\label{ExpAdvImgs-FGSM-S}]{\includegraphics[scale=0.5]{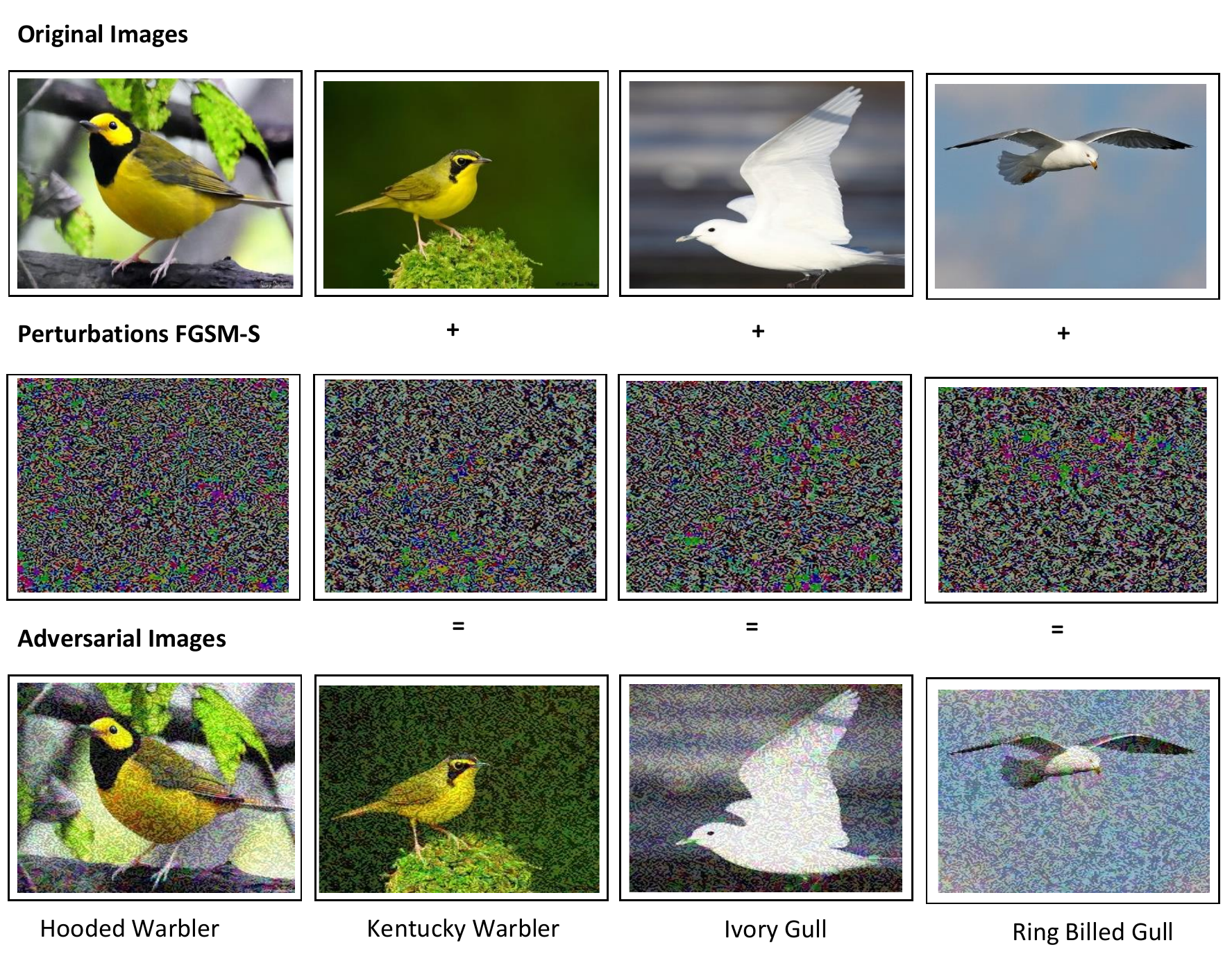}}
        \subfloat[\vspace{-1em}Itr-S based adversarial images\label{ExpAdvImgs-Itr-S}]{\includegraphics[scale=0.5]{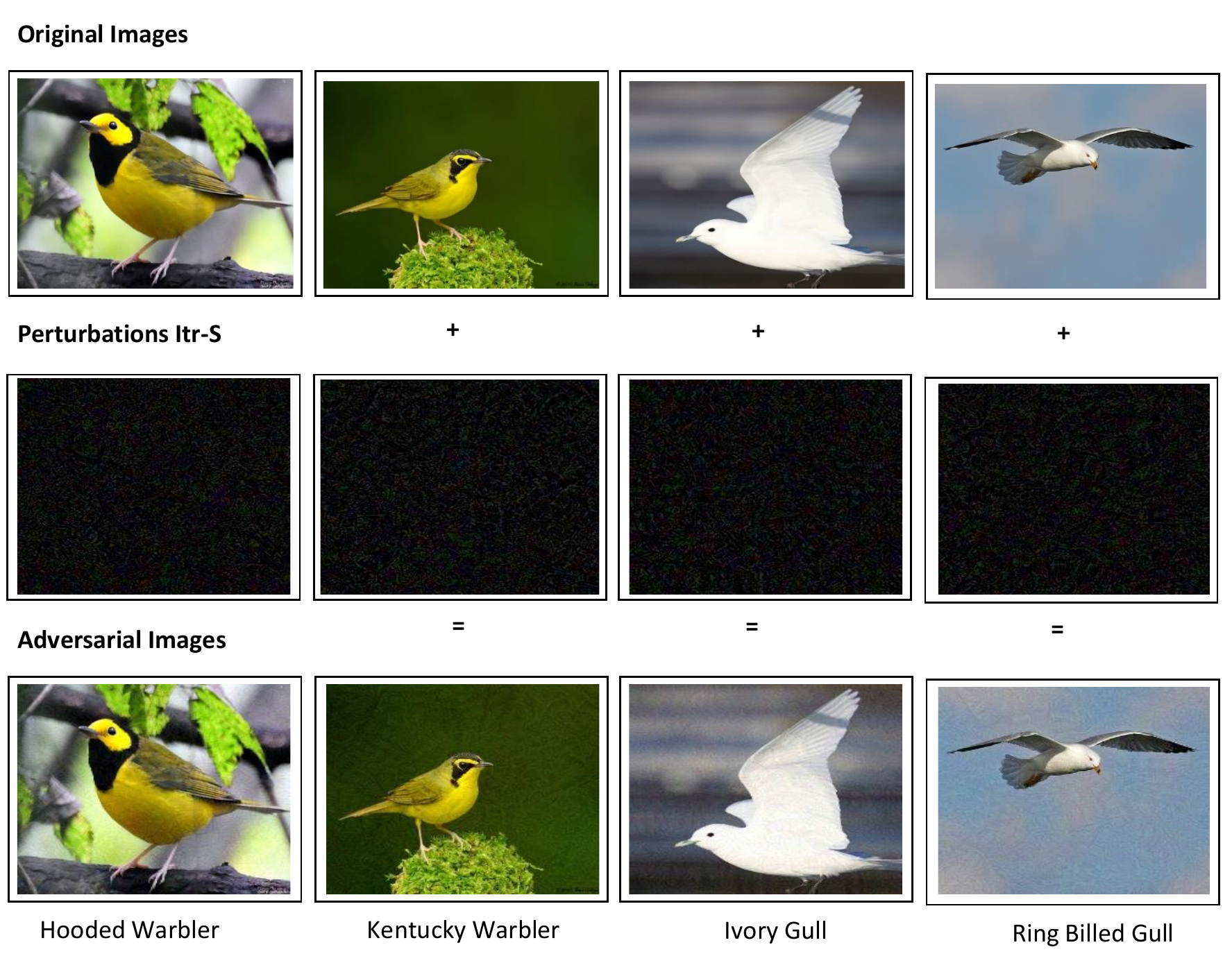}}
        
	\vspace{1.5em}
	\caption{Example bird images of four different species. Original images are in the first row. The relevant adversarial perturbations are in the second row. The resultant adversarial images are in the third row.}
	\label{ExpAdvImgs-FGSM}
\end{figure*}

\subsection{Experiments}
\label{ExpsMulti}
For all the experiments, the $10$-fold cross-validation technique is applied. The adversarial attacks are applied to the testing images only. The effectiveness of the novel system is evaluated both on the original fold-test images and then on the adversarial images.

Three variants of SIFT features are computed by utilizing the following parameter values: patchSize: $64$, $128$, and $256$; maxBinValue: $0.2$, numOrientationBins: $8$, and numSpatialBins: $2$. Similarly, three variants of HOG features are computed by utilizing the following parameter values: image size: $64, 126$, and $256$; pixels per cell: $32, 64$, and $128$. These parameter values help us to get a variety of SIFT and HOG features. 

Two variants of FGSM attacks are applied to the testing images to generate adversarial images, i.e. medium-level attack FGSM-M (epsilon (max norm) value $50$) and strong attack FGSM-S (epsilon $150$) \cite{advAttackGens}. Moreover, two variants of Iterative attacks are applied to the testing images to generate adversarial images, i.e. medium-level attack Itr-M (epsilon $18$, alpha $1$, and iterations $10$), and strong attack Itr-S (epsilon $50$, alpha $1$, and iterations $10$) \cite{advAttackGens}. Sample intact and adversarial images generated by applying FGSM-M, FGSM-S, Itr-S, and Itr-M are shown in Figs. \ref{ExpAdvImgs-FGSM-M}, and \ref{ExpAdvImgs-FGSM-S}, \ref{ExpAdvImgs-Itr-M}, and \ref{ExpAdvImgs-Itr-S} respectively. The experimental results of the novel system are compared with the results generated from four state-of-the-art deep models that have been commonly used in image classification tasks, i.e. ResNet \cite{he2016deep}, AlexNet \cite{krizhevsky2012imagenet}, VGG \cite{simonyan2014very}, and SqueezeNet \cite{iandola2016squeezenet}.

The experimental results demonstrate that the lateralized approach is scalable to multi-class tasks, see Table \ref{ExpResTableMulti}. The novel lateralized system (LatSys) outperformed all the state-of-the-art deep models for the classification of original test images by $19.05\%-41.02\%$. Moreover, the novel system successfully exhibited robustness against three types of adversarial attacks (FGSM-M, Itr-M, and Itr-S). For all these attacks, the classification accuracy of the novel lateralized system is between $19.05\%-41.02\%$ and $61.17\%$, whereas, the classification accuracy of all other systems is between  $05.03\%-41.02\%$ and $42.03\%$. The novel system could not completely resist the FGSM-S attack because it is a very strong adversarial attack that destroys the image contents badly. But all other deep models perform worse. The novel lateralized system outperformed all the state-of-the-art deep models for the classification of adversarial images by $1.36\%-49.22\%$.

\begin{table}[!t]
	\caption{Classification Accuracy \\ 
            \scriptsize Highest Accuracy is in bold.}
	\label{ExpResTableMulti}
	\resizebox{\columnwidth}{!}{%
	\begin{tabular}{|c|c|c|c|c|c|}
        \hline
        \textbf{}         & \textbf{VGG} & \textbf{SqueezeNet} & \textbf{AlexNet} & \textbf{ResNet} & \textbf{LatSys} \\ \hline
        \textbf{OrigImgs} & 50.18        & 54.78               & 35.38            & 57.35           & \textbf{76.40}               \\ \hline
        \textbf{FGSM-M}   & 15.83        & 19.40               & 15.85            & 15.08           & \textbf{41.28}               \\ \hline
        \textbf{FGSM-S}   & 00.99        & 03.07               & 01.60            & 02.61           & \textbf{04.43}               \\ \hline
        \textbf{Itr-M}    & 42.03        & 37.05               & 28.99            & 13.06           & \textbf{61.17}               \\ \hline
        \textbf{Itr-S}    & 31.08        & 28.58               & 24.65            & 05.03           & \textbf{54.25}               \\ \hline
        \end{tabular}
	}
\end{table}

\subsection{Interpretation of Decisions}
\label{IntDec}
The decision-making procedure of the novel lateralized system is interpretable. The role played by various system components is illustrated by using four cases. Each case is further explained by utilizing two examples. Those examples are chosen when the system's components are not consistent with one another, i.e., all components do not agree on the correct class. However, the final prediction of the lateralized system is correct.

\subsubsection{Case-1}
\begin{figure}
	\begin{center}
		\includegraphics [scale=0.49]{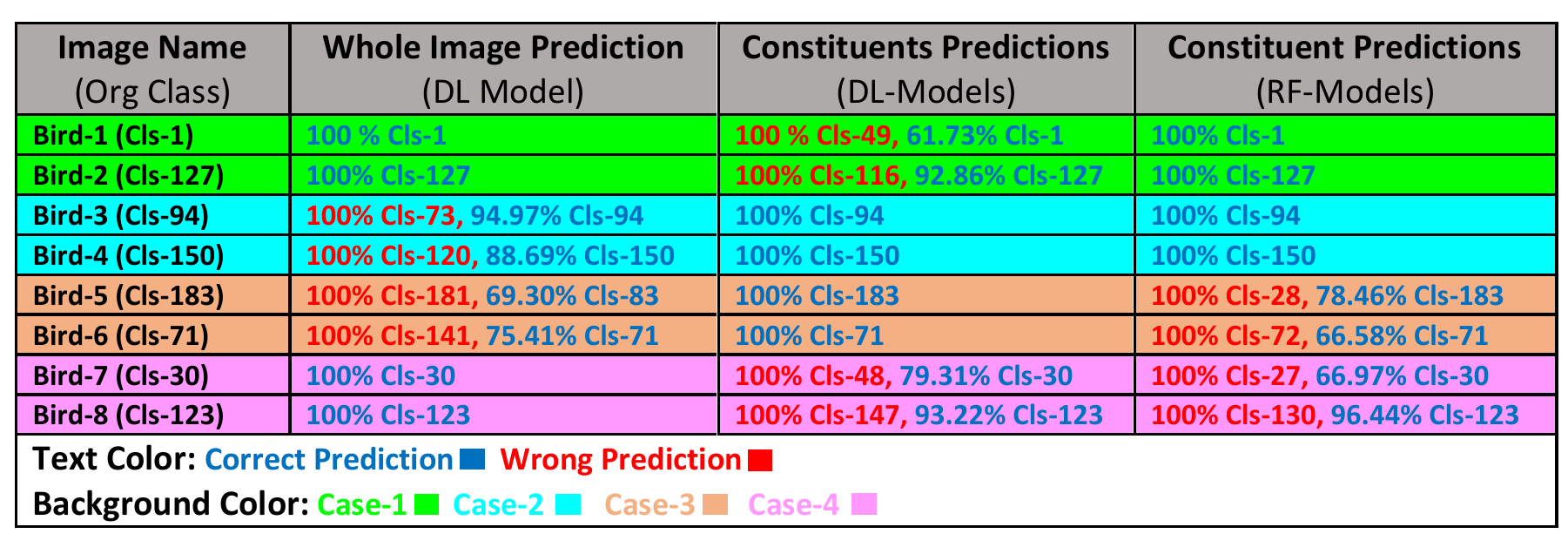}
		\caption{Interpretation of decision-making process adopted by the lateralized system to classify adversarial images.}
		\label{InterpRes}
	\end{center}
\end{figure}

In this case, the holistic-level prediction made by the deep model for the whole image is correct. Whereas the constituent-level prediction made by the component-level deep models is incorrect, i.e. the $ \mathcal{CLP}$ and $ \mathcal{HLP}$ are at odds with each other. However, the constituent-level prediction made by the RF models is correct, which supports the holistic-level prediction. Consequently, the final prediction made by the lateralized system is correct. For example, during the prediction of the image Bird-1 (black-footed albatross, \textit{class-1}), the holistic-level deep model predicts that it is a $100\%$ \textit{class-1}. Whereas, the constituent-level deep models predict that it is a $100\%$ \textit{class-49} and $61.73\%$ \textit{class-1}. However, the RF models predict that it is a $100\%$ \textit{class-1}. This prediction class supports the holistic-level prediction class. Consequently, it enables the lateralized system to make a correct final prediction for the Bird-1, i.e. \textit{class-1}, see Fig \ref{InterpRes}. Similarly, during the prediction of the image Bird-2 (Savannah sparrow, \textit{class-127}), the holistic-level deep model predicts that it is a $100\%$ \textit{class-127}. Whereas, the constituent-level deep models predict that it is a $100\%$ \textit{class-116} and $92.86\%$ \textit{class-127}. However, the RF models predict that it is a $100\%$ \textit{class-127}. This prediction class supports the holistic-level prediction class. Consequently, it enables the lateralized system to make a correct final prediction for the Bird-2, i.e. \textit{class-127}, see Fig \ref{InterpRes}.

\subsubsection{Case-2}
In this case, the holistic-level prediction made by the deep model for the whole image is incorrect. Whereas the constituent-level prediction made by the component-level deep models is correct, i.e. the $ \mathcal{CLP}$ and $ \mathcal{HLP}$ are at odds with each other. However, the constituent-level prediction made by the RF models is correct, which supports the constituent-level deep models' prediction. Consequently, the final prediction made by the lateralized system is correct. For example, during the prediction of the image Bird-3 (white-breasted nuthatch, \textit{class-94}), the holistic-level deep model predicts that it is a $100\%$ \textit{class-73} and $94.97\%$ \textit{class-94}. Whereas, the constituent-level deep models predict that it is a $100\%$ \textit{class-94}. However, the RF models predict that it is a $100\%$ \textit{class-94}. This prediction class supports the constituent-level deep models' prediction class. Consequently, it enables the lateralized system to make a correct final prediction for the Bird-3, i.e. \textit{class-94}, see Fig \ref{InterpRes}. Similarly, during the prediction of the image Bird-4 (sage thrasher, \textit{class-150}), the holistic-level deep model predicts that it is a $100\%$ \textit{class-120} and $88.69\%$ \textit{class-150}. Whereas, the constituent-level deep models predict that it is a $100\%$ \textit{class-150}. However, the RF models predict that it is a $100\%$ \textit{class-150}. This prediction class supports the constituent-level deep models' prediction class. Consequently, it enables the lateralized system to make a correct final prediction for the Bird-4, i.e. \textit{class-150}, see Fig \ref{InterpRes}.

\subsubsection{Case-3}
In this case, the holistic-level prediction made by the deep model for the whole image is incorrect. Whereas the constituent-level prediction made by the component-level deep models is correct, i.e. the $ \mathcal{CLP}$ and $ \mathcal{HLP}$ are at odds with each other. Subsequently, the constituent-level prediction made by the RF models is also incorrect. But the prediction values for the correct class generated by the holistic-level deep model and constituent-level RF models assist the lateralized system to make a final correct prediction. For example, during the prediction of the image Bird-5 (Northern water thrush, \textit{class-183}), the holistic-level deep model predicts that it is a $100\%$ \textit{class-181} and $69.30\%$ \textit{class-183}. Whereas, the constituent-level deep models predict that it is a $100\%$ \textit{class-183}. Subsequently, the RF models predict that it is a $100\%$ \textit{class-28} and $78.46\%$ \textit{class-183}. These prediction values for the correct \textit{class-183} assist the lateralized system to make a correct final prediction for the Bird-5, i.e. \textit{class-183}, see Fig \ref{InterpRes}. Similarly, during the prediction of the image Bird-6 (long-tailed Jaeger, \textit{class-71}), the holistic-level deep model predicts that it is a $100\%$ \textit{class-141} and $75.41\%$ \textit{class-71}. Whereas, the constituent-level deep models predict that it is a $100\%$ \textit{class-71}. Subsequently, the RF models predict that it is a $100\%$ \textit{class-72} and $66.58\%$ \textit{class-71}. These prediction values for the correct \textit{class-71} assist the lateralized system to make a correct final prediction for the Bird-6, i.e. \textit{class-71}, see Fig \ref{InterpRes}.

\subsubsection{Case-4}
In this case, the holistic-level prediction made by the deep model for the whole image is correct. Whereas the constituent-level prediction made by the component-level deep models is incorrect, i.e. the $ \mathcal{CLP}$ and $ \mathcal{HLP}$ are at odds with each other. Subsequently, the constituent-level prediction made by the RF models is also incorrect. But the prediction values for the correct class generated by the constituent-level deep model and RF models assist the lateralized system to make a final correct prediction. For example, during the prediction of the image Bird-7 (fish crow, \textit{class-30}), the holistic-level deep model predicts that it is a $100\%$ \textit{class-30}. Whereas, the constituent-level deep models predict that it is a $100\%$ \textit{class-48} and $79.31\%$ \textit{class-30}. Subsequently, the RF models predict that it is a $100\%$ \textit{class-27} and $66.97\%$ \textit{class-30}. These prediction values for the correct \textit{class-30} assist the lateralized system to make a correct final prediction for the Bird-7, i.e. \textit{class-30}, see Fig \ref{InterpRes}. Similarly, during the prediction of the image Bird-8 (Henslow sparrow, \textit{class-71}), the holistic-level deep model predicts that it is a $100\%$ \textit{class-123}. Whereas, the constituent-level deep models predict that it is a $100\%$ \textit{class-147} and $93.22\%$ \textit{class-123}. Subsequently, the RF models predict that it is a $100\%$ \textit{class-130} and $96.44\%$ \textit{class-123}. These prediction values for the correct \textit{class-123} assist the lateralized system to make a correct final prediction for the Bird-8, i.e. \textit{class-123}, see Fig \ref{InterpRes}.

\section{Discussion}
\label{disc}
LatSys was designed to provide robust solutions for real-world problems that include uncertainty, noise, and irrelevant and redundant data. It is noted that this work aimed to create a system that exhibits natural robustness against adversarial attacks and not to devise another adversarial avoidance technique for a specific model or specific adversarial attack. Although the novel architecture was not designed to model human (or other animals) vision, vertebrate brains do have complementary lateralized modules that represent objects at local (left) and global (right) levels. 

LatSys has the ability to simultaneously represent the same problem instance at different levels of abstraction. This enabled the novel system to make correct predictions for the images that are badly corrupted by adversarial attacks. Although the system's components were not consistent with one another while predicting those images, i.e., all components did not agree on the correct class. But the final prediction made by the LatSys was correct (see Section \ref{IntDec}. This ability to tackle the problem at different levels enables the novel system to successfully demonstrate robustness against different types of adversarial attacks. An adversarial attack must successfully challenge both the holistic components and constituents of an image in order to deceive the novel system.

It has been reported that built-in support for heterogeneity and niche-based algorithm of LCSs play a critical role in developing lateralized systems \cite{siddique2020learning,siddique2020lateralized,siddique2021frames,siddique2022lateralized}. However, the improved version of the lateralized system shows that the lateralized approach is scalable and not limited to LCSs. Instead of LCSs, it is the lateralized architecture that provides robustness against adversarial attacks. In this work, RF models make linearity assumptions to handle a large number of classes practically.

The decision-making procedure of the novel lateralized system is interpretable, see Section \ref{IntDec}. This is a step toward eXplainable AI \cite{gunning2019xai,adadi2018peeking}. The novel system looks like an ensemble system \cite{zhang2012ensemble} because it solves different components of the problem to take a final decision. However, the use of excite and inhibit signals to activate and deactivate different system modules and the ability to consider the same problem component (e.g. face) at different levels of abstraction makes it a lateralized system rather than an ensemble system.

A lateralized system may work similarly to a conventional AI system if the constituent knowledge is not available or is not reused to solve higher-level problem components. Moreover, this work is based on continuous/transfer learning and does not achieve end-to-end learning. Although it is expensive to learn constituent knowledge, once learned, it can be reused to correctly identify the class of a noisy or adversarial image, even if the image as a whole, except a part, is corrupted. Moreover, this acquired knowledge can be reused at a constituent level by the lateralized system for other problems, for example, headlights in a car and eyes in a face can be reused for the classification of highway congestion against a road mass.

\section{Conclusion}
\label{Con}
The novel system successfully applied lateralization and modular learning at multiple levels of abstraction to show the effectiveness, scalability, and robustness of the lateralized approach. The constituents class matrix allows the lateralized system to successfully handle multiple classes. The methodologies of constituent-level perception and holistic-level perception enable the novel system to correctly classify corrupted images. The heterogeneous features-based strategy facilitated the representation of holistic patterns of a bird as well as its constituent parts such as the beak, neck, wing, etc. The learned patterns at constituent levels are successfully utilized at the holistic level to aid classification. Inhibit and excite signals are generated based on the feedback from the context phase. Inhibit signals are used to avoid extraneous computations while addressing simple images, whereas, excite signals are used to give more attention to noisy and corrupt images. This strategy enables the novel system to generate correct final predictions for badly corrupted images where the holistic-level predictions and constituent-level predictions are at odds with each other. The experimental results show that the novel lateralized system outperformed all the state-of-the-art deep learning-based systems for the classification of normal and adversarial images by $19.05\%-41.02\%$ and $1.36\%-49.22\%$, respectively. Despite the promising results, the novel system depends on constituent-level labeled data sets. Future work will consider how constituent parts can be learned in an end-to-end or continual learning manner. 

\section*{Acknowledgment}
This work is supported by Science for Technological Innovation National Science Challenge, New Zealand.

\bibliographystyle{IEEEtran}

\bibliography{BibAB4Trans}

\ifCLASSOPTIONcaptionsoff
  \newpage
\fi

%
\vspace{-2.5em}
\begin{IEEEbiography}[{\includegraphics[width=1in,height=1.25in,clip,keepaspectratio]{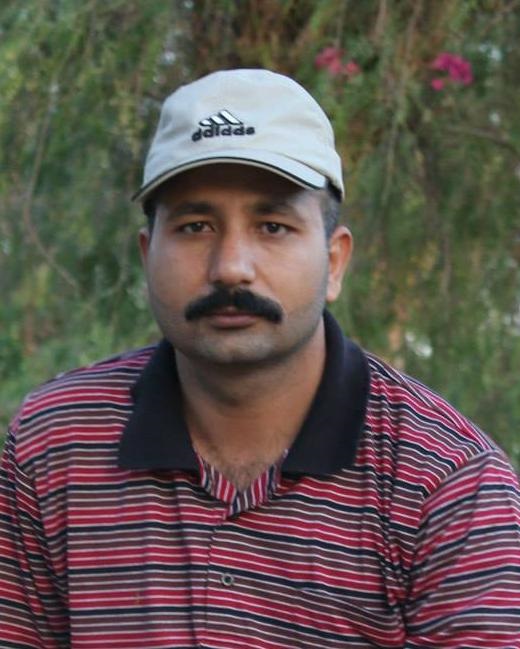}}]{Abubakar Siddique} is a PhD student at School of Engineering and Computer Science, Victoria University of Wellington. His research interests are Learning Classifier Systems, Evolutionary Computing, Brain Lateralization, and Cognitive Neuroscience. Mr Siddique did his undergraduate (major in Computer Science) from Quaid-i-Azam University Islamabad and Master in Computer Engineering from U.E.T Taxila. He was the recipient of ``Student Of The Session'' award. His undergraduate senior project was conducted in an internship at Ultimus, where his work was deployed at the company's Workflow product. He spent nine years at Elixir Pakistan, a California based software Company. His last designation was a Principal Software Engineer where he lead a team of software developers. He developed enterprise level software for customers such as Xerox, IBM and Finis.
\end{IEEEbiography}
\vspace{-2.5em}
\begin{IEEEbiography}[{\includegraphics[width=1in,height=1.25in,clip,keepaspectratio]{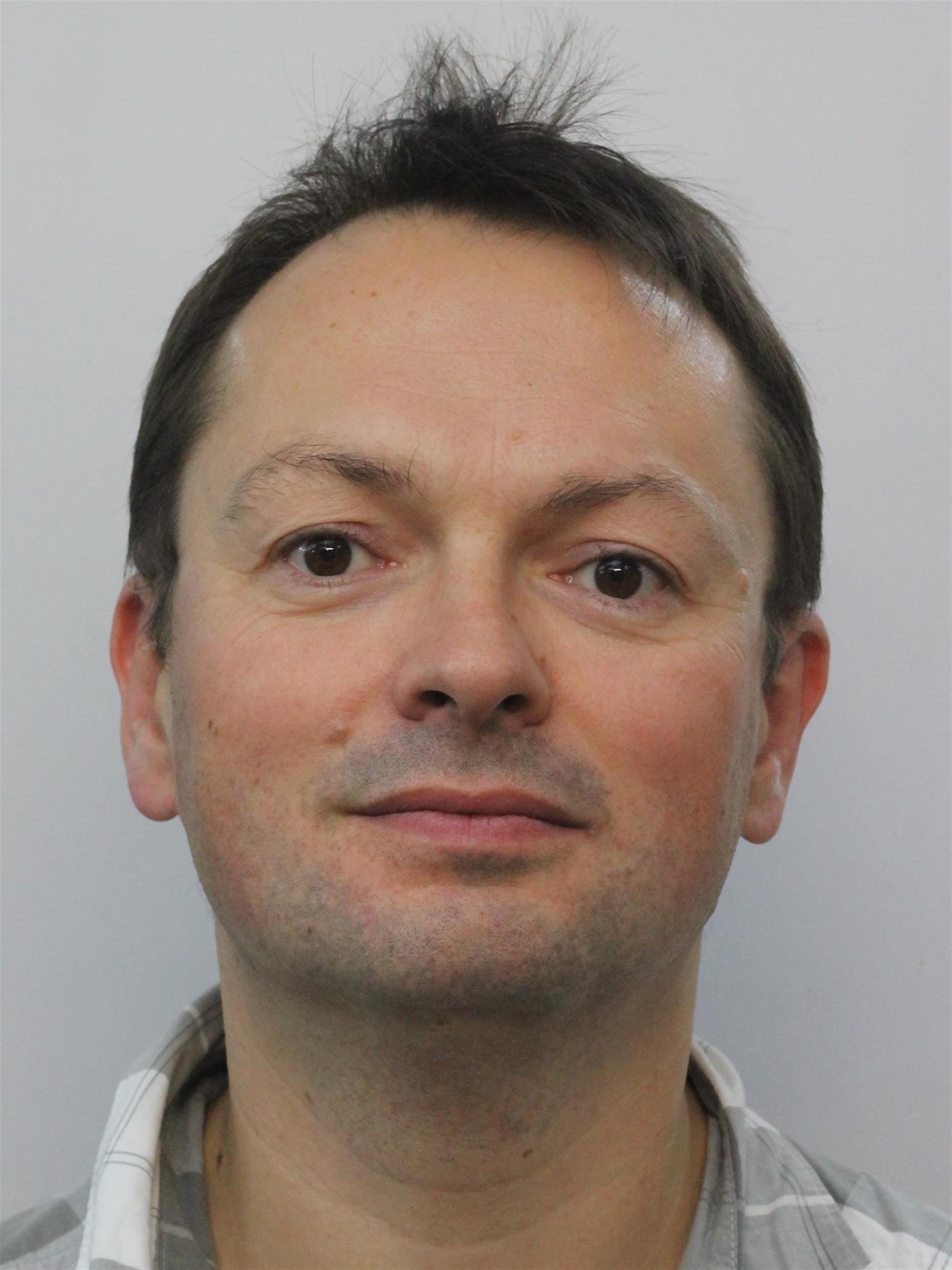}}]{Will N. Browne's} research interests are in developing Artificial Cognitive Systems. His background is in Mechanical Engineering (B.Eng. degree (Hons.)) from the University of Bath, U.K., 1993, and both the M.Sc. degree in Energy and the EngD degree (Engineering Doctorate Scheme) from the University of Wales, Cardiff, U.K., in 1994 and 1999, respectively. After lecturing for eight years at the Department of Cybernetics, University of Reading, Reading, U.K., he is now an Associate Professor with the Evolutionary Computation Research Group, School of Engineering and Computer Science, Victoria University of Wellington, New Zealand. He has published over 100 academic papers in books, refereed international journals, and conferences. This includes two best paper awards at the ACM Genetic and Evolutionary Computation Conference (GECCO), where he has also served as a track chair on four occasions in the Evolutionary Machine Learning Track (or equivalent). He serves on the editorial board of Applied Soft Computing Journal. Together with Dr Ryan Urbanowicz, he has authored the textbook ‘Introduction to Learning Classifier Systems’, Springer, 2017. He was the Co-Local Chair for the IEEE Congress Evolutionary Computation (CEC), Wellington, 2019.
\end{IEEEbiography}
\vspace{-2.5em}
\begin{IEEEbiography}[{\includegraphics[width=1in,height=1.25in,clip,keepaspectratio]{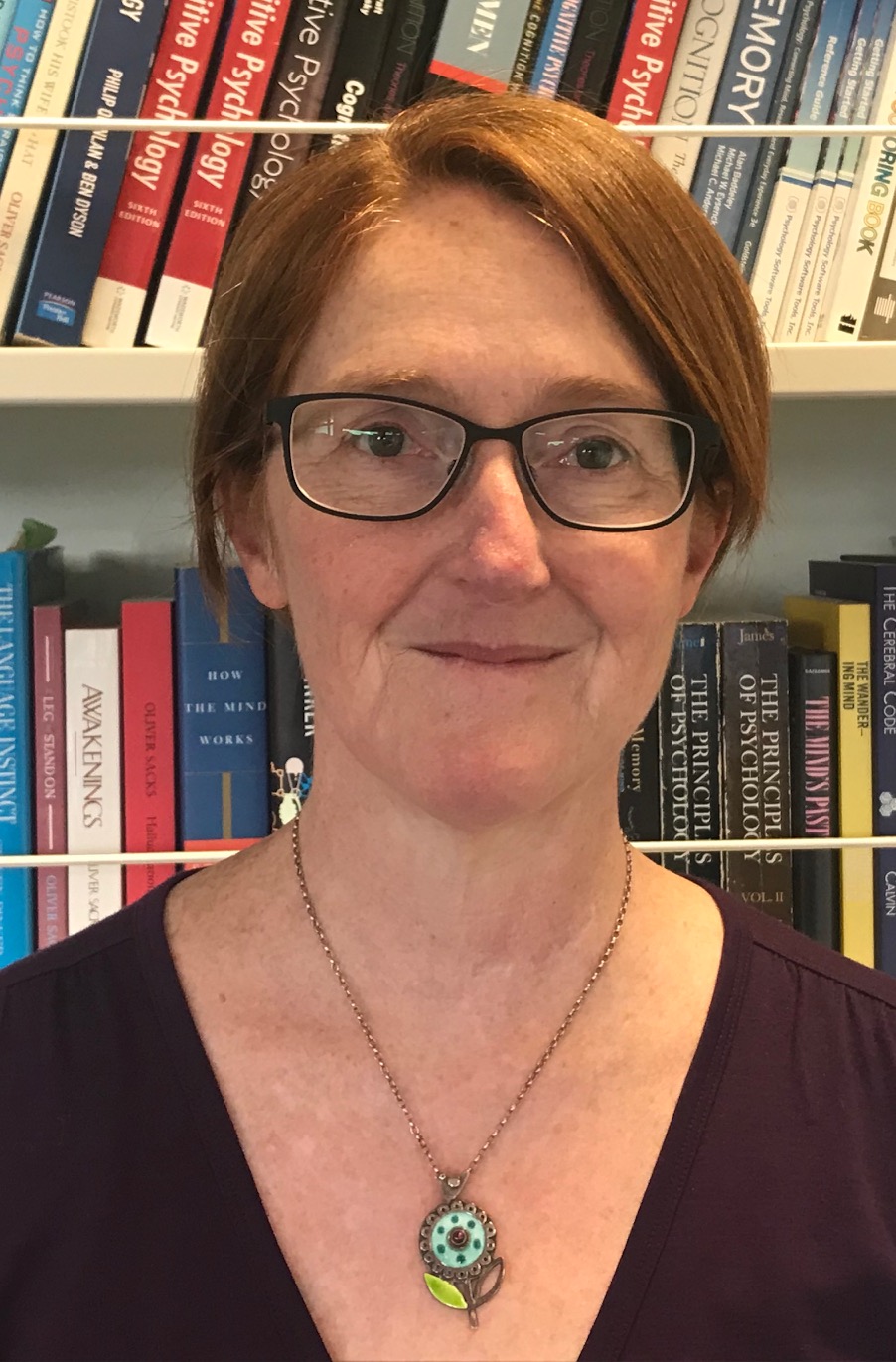}}]{Gina M. Grimshaw} received a BSc in Biochemistry from the University of Toronto in 1987, and a PhD in Cognitive Psychology from the University of Waterloo in 1996. She was a post-doctoral fellow in the Department of Cognitive Science at the University of California San Diego (1996-1997) before taking up an academic position at California State University San Marcos. Since 2007 she has been at Victoria University of Wellington, where she is Associate Professor of Psychology, Programme Director of Cognitive and Behavioural Neuroscience, and Director of the Cognitive and Affective Neuroscience Lab. Her research explores the cognitive and neural mechanisms that support cognition and emotion, with a particular focus on hemispheric specialization and interaction. She has supervised over 20 postgraduate students in Psychology, Cognitive and Behavioural Neuroscience, and Engineering. Her research has been funded by the National Institute of Mental Health (US) and the Royal Society of New Zealand Marsden Fund. She has authored over 50 refereed journal publications, and is Editor of Laterality: Asymmetries of Body, Brain, and Cognition (2016 – present). She is Secretary of the Australasian Society for Experimental Psychology, and chaired the Society’s Experimental Psychology Conference (EPC) in 2011 and 2019. She has won university awards for Teaching Excellence, Research Excellence, and Contributions to Equity and Diversity. 
\end{IEEEbiography}




\end{document}